\documentclass[journal]{IEEEtran}

\usepackage{booktabs}
\usepackage{graphicx}
\usepackage{textcomp}
\usepackage{xcolor}
\usepackage{cite}
\usepackage{amsmath,amssymb,amsfonts,bm}

\usepackage{algorithm}
\usepackage{algorithmic}

\usepackage{makecell}
\usepackage{multirow}
\usepackage{adjustbox}
\usepackage{amsthm}
\usepackage{tabularx}
\usepackage{array}
\usepackage{caption}
\usepackage{subcaption}

\usepackage[colorlinks=true,urlcolor=blue]{hyperref}
\begin{document}
%
\title{From Shared Demand Patterns to Local Uncertainty: Probabilistic Load Forecasting by Mixing Compact Adaptations}
%
%
%

\author{Haoran~Li,~\IEEEmembership{Member,~IEEE,}
Zhe~Cheng,~\IEEEmembership{Student Member,~IEEE,}
Yang~Weng,~\IEEEmembership{Senior Member,~IEEE,}
\vspace{-10mm}
}

\maketitle

\begin{abstract}

Probabilistic load forecasting has been widely studied for power-system operation and planning, but customer- and transformer-level forecasting introduces a distinct scalability challenge. At these levels, load uncertainty is strongly affected by customer behavior, weather, and mixed load composition, making it difficult for a single shared model to capture heterogeneous patterns. Using separate probabilistic models can improve local accuracy, but becomes costly to train, store, update, and validate at scale. To address this challenge, we develop a scalable customer-aware forecasting framework that learns common demand behavior through a shared model while adapting only a compact subset of parameters. Rather than using an independent model for each load or assigning each load to a specialized model, the proposed design learns a small bank of low-dimensional adaptation components and allows each load to combine them according to its forecasting characteristics. This preserves shared knowledge across customers while providing sufficient flexibility for heterogeneous and mixed load compositions. Experiments on 590 load profiles from the SMART-DS dataset show consistent improvements in deterministic accuracy and probabilistic quality over statistical, neural-network, Transformer-based, and pretrained time-series baselines, while retaining low storage and inference costs. The source code and data processing pipeline are available at \href{https://github.com/haorandd/M2oE2_load_forecast.git}{\textcolor{blue}{M$^2$oE$^2$-mLoRA}}.
\end{abstract}


%
\IEEEpeerreviewmaketitle

\vspace{-2mm}
\section{Introduction}
\vspace{-1mm}

Load forecasting is a foundational function for modern power systems, providing essential information for economic dispatch~\cite{han2021task}, demand response~\cite{pramono2019deep}, reliability assessment~\cite{saxena2024intelligent}, event detection~\cite{ref:Haoran2022D,ref:Haoran2022T,ref:Haoran2023S}, and infrastructure planning~\cite{zhang2023novel}. As smart meters and advanced sensing infrastructures become increasingly available, forecasting practice is moving beyond aggregate system-level prediction toward fine-grained, bottom-up forecasting for individual customers in distribution feeders~\cite{ref:2019ChengjinD,li2025exarnn,li2025external}. This shift is particularly important for distribution-system operation and can support distributed energy resource coordination~\cite{attarha2019affinely}, demand-side management~\cite{kong2019improved}, local market participation~\cite{schreck2020methodological}, and capacity planning~\cite{ref:Ponoćko2018F}. Compared with point forecasting, probabilistic load forecasting further characterizes predictive uncertainty and provides richer information for risk-aware decision-making under variable and disaggregated demand~\cite{wang2022novel}. 

Nevertheless, customer- and transformer-level probabilistic forecasting remains highly challenging in realistic distribution feeders with massive heterogeneous loads, including residential, commercial, industrial, and mixed transformer-level demand. Unlike more aggregated load forecasting, local power measurements reflect the combined effects of largely unobserved factors such as occupancy, business activity, appliance usage, electric-vehicle charging, distributed energy resources, and customer composition behind transformers. These factors can produce strong differences in consumption and uncertainty patterns across customers, phases, locations, and aggregation levels, making it difficult for a single global forecaster to fully capture local behavior~\cite{wang2018combining,10138375}. At the same time, training and maintaining an independent deep probabilistic model for each customer or transformer is impractical for utilities, as storage, computation, and model-management costs grow rapidly with the number of loads. Therefore, scalable forecasting requires a mechanism that shares common knowledge across massive customers while adapting compactly to customer-specific demand and uncertainty characteristics.

Early approaches address local heterogeneity by building separate forecasting models for different customers, buildings, or load groups. Clustering-based methods partition customers according to load profiles and train separate predictors for the resulting clusters~\cite{wijaya2015cluster,fahiman2017improving}. Transfer-learning-based methods instead transfer knowledge learned from data-rich source loads, buildings, or customer groups to related target domains with limited data~\cite{ribeiro2018transfer,wu2022efficient,tian2019similarity}. While clustering can substantially reduce complexity compared with maintaining an independent forecaster for every customer, storage and model-management costs may still become considerable when many clusters are required to represent heterogeneous loads. Transfer learning improves knowledge reuse, but often relies on source-to-target transfer and repeated adaptation across different loads.

The third category follows a shared-model adaptation strategy, where knowledge is learned jointly from many related loads and then adapted to specific customers or customer groups. Meta-learning methods learn a shared initialization or meta-model that can be quickly adapted to target household loads using limited local data~\cite{he2022transferrable,tsoumplekas2025few}, but typically require an additional learning-to-adapt optimization procedure. Personalized federated learning trains a common model across consumers while allowing local parameter or gradient updates to preserve customer-specific characteristics~\cite{wang2022personalized,briggs2022federated,he2023privacy}. Hypernetwork-based methods instead generate customer-specific model parameters or compact sub-parameters conditioned on customer information~\cite{danish2024leveraging,li2025external}, while fine-tuning approaches adapt only selected parameters of a pretrained model for different customer types~\cite{gupta2024low}. These approaches substantially improve scalability by sharing most model knowledge across loads, but still face a capacity-efficiency tradeoff: the adapted parameter space must remain compact for large-scale deployment while being expressive enough to capture heterogeneous customer behavior and mixed transformer-level load compositions.

To address the capacity-efficiency challenge, we first investigate where customer-specific adaptation should be introduced within a shared probabilistic forecaster. A key observation from our parameter-selection and ablation tests is that the forecasting model can be naturally separated into two functional stages: learning generalizable temporal and contextual representations across customers, and adapting these shared representations to customer-specific demand and uncertainty patterns. While the former can be largely shared, different customers may require different mappings from the common representation to their predictive distributions. This suggests that customer-specific adaptation should be concentrated at the post-fusion prediction stage, identifying the probabilistic output head as the key parameter subset for adaptation.

We therefore introduce low-rank adaptation (LoRA)~\cite{hu2022lora} into the output head, freezing the shared parameters while learning compact low-rank updates for different load groups. This preserves the common output-mapping structure while substantially reducing the number of adapted parameters. However, a single LoRA update restricts adaptation and may not provide sufficient flexibility for heterogeneous customers and mixed transformer-level loads; simply increasing its rank improves capacity but also increases parameter cost. We thus propose a mixture-of-LoRA mechanism that represents each adaptation update as a weighted combination of multiple low-rank components, allowing diverse adaptation patterns to be composed from a small reusable parameter bank. The shared probabilistic forecaster is implemented using M$^2$oE$^2$~\cite{li2025external}, which incorporates external information as meta-knowledge and has demonstrated strong performance on benchmarks including GEFCom2014~\cite{hong2016probabilistic}.

The resulting model is termed M$^2$oE$^2$-mLoRA. We evaluate it on 590 residential, commercial, and industrial load profiles from the SMART-DS dataset~\cite{smartds2020} against statistical models, recurrent and convolutional neural networks, Transformers, and pretrained time-series foundation models. The evaluation covers deterministic accuracy, probabilistic forecasting quality, inference time, and storage cost. Results show that M$^2$oE$^2$-mLoRA improves both deterministic and probabilistic forecasting performance while maintaining low computational and storage overhead, demonstrating a favorable accuracy--uncertainty--cost tradeoff for customer- and transformer-level forecasting.




The rest of the paper is as follows. Section \ref{sec:prob} formulates the problem.
Section \ref{sec:pre} provides preliminaries of M$^2$oE$^2$. Section \ref{sec:model} illustrates M$^2$oE$^2$-mLoRA framework. Section \ref{sec:exp} presents numerical results, and Section \ref{sec:con} concludes the paper.

\vspace{-3mm}
\section{Problem Formulation}
\vspace{-1mm}

\label{sec:prob}

We define the customer-aware probabilistic load forecasting problem for massive heterogeneous feeder loads as follows.

\begin{itemize}
\item Goal: Learn a scalable probabilistic forecasting model that shares common knowledge across massive customer- and transformer-level loads, while adapting compactly to different load groups in realistic distribution feeders.

\item Given: A customer/transformer index set $\mathcal{C}$, where each $c\in\mathcal{C}$ denotes one customer-level or transformer-level load. The historical load measurements of $c$ are denoted as $\{x_{c,i}\}_{i=1}^{N}$, where $N$ is the number of measurement points. Optional external measurements are denoted as $\{w_{j,i}\mid j=1,\cdots,M; i=1,\cdots,N\}$, where $M$ is the number of external data sources, such as weather, calendar variables, and electricity prices. The set $\mathcal{C}$ may contain $K$ load groups, including residential, commercial, industrial, and transformer-level loads with different mixed customer compositions.

\item Find: A shared M$^2$oE$^2$ probabilistic forecaster $f_{\Theta}$ and a set of adaptation parameters $\{\theta_k\}_{k=1}^{K}$, where $\Theta$ denotes the shared global model parameters and $\theta_k$ denotes the LoRA parameters associated with the $k$-th load cluster.
\end{itemize}

\vspace{-3mm}
\section{Preliminary:
\texorpdfstring{M$^2$oE$^2$}{M2oE2}
Global Forecaster}
\vspace{-1mm}

\label{sec:pre}

This section briefly reviews our previous M$^2$oE$^2$ probabilistic forecaster~\cite{li2025external}, which serves as the shared global forecasting backbone $f_{\Theta}$ in this paper. The model is applicable to each customer or transformer $c\in\mathcal{C}$ and is trained using load data from all $c\in\mathcal{C}$. For notational simplicity, we omit the customer/transformer subscript $c$ in this section. M$^2$oE$^2$ builds on a probabilistic sequence base model and further treats external data as meta-knowledge to guide feature extraction. It uses MetaNets to transform heterogeneous external sources into candidate experts, a sparse gating network to select informative experts, and a residual anchor to preserve the no-external-data case. For more details, please review \cite{li2025external}. 

\vspace{-3mm}
\subsection{Variational Sequence As A Base Model}
\vspace{-1mm}

\label{subsec:pre_base}

Let $\{x_i\}_{i=1}^{N}$ denote a load measurement sequence, where $x_i$ is a scalar load measurement. A sequence base model, such as RNN, GRU, or LSTM, first encodes historical load information into a hidden state $\boldsymbol{h}_i$. For example, an RNN-type base model updates the hidden state as
\begin{equation}
\label{eqn:pre_rnn}
\boldsymbol{h}_i
=
\rho\left(W_{hx}x_i+W_{hh}\boldsymbol{h}_{i-1}+\boldsymbol{b}_h\right),
\end{equation}
where $\rho(\cdot)$ is the activation function, and $W_{hx}$, $W_{hh}$, and $\boldsymbol{b}_h$ are trainable parameters. The hidden state $\boldsymbol{h}_i$ summarizes historical load patterns and provides the temporal feature for future forecasting.

To quantify uncertainty, M$^2$oE$^2$ adopts a variational probabilistic forecasting structure on top of the sequence base model. A latent vector $\boldsymbol{z}_{i+1}$ is introduced to represent stochastic factors that affect the future load. The encoder approximates the posterior distribution of $\boldsymbol{z}_{i+1}$ as
\begin{equation}
\label{eqn:pre_encoder}
\begin{aligned}
&q_{\Theta}\left(
\boldsymbol{z}_{i+1}\mid x_i,\boldsymbol{h}_{i-1}
\right)\\
&\quad=\mathcal{N}\Bigl(
\boldsymbol{\mu}_z(x_i,\boldsymbol{h}_{i-1}),
\operatorname{diag}\bigl(
\boldsymbol{\sigma}_z^2(x_i,\boldsymbol{h}_{i-1})
\bigr)\Bigr).
\end{aligned}
\end{equation}
where $\boldsymbol{\mu}_z(\cdot)$ and $\boldsymbol{\sigma}_z^2(\cdot)$ are neural-network mappings. The decoder then predicts the mean and variance of the future load distribution:
\begin{equation}
\label{eqn:pre_decoder}
p_{\Theta}\left(x_{i+1}\mid \boldsymbol{z}_{i+1}\right)
=
\mathcal{N}\left(
\mu_x\left(\boldsymbol{z}_{i+1}\right),
\sigma_x^2\left(\boldsymbol{z}_{i+1}\right)
\right).
\end{equation}
The model is trained by minimizing the negative evidence lower bound:
\begin{equation}
\label{eqn:pre_elbo}
\begin{aligned}
\mathcal{L}_{\mathrm{ELBO}}(\Theta)
=
&-\mathbb{E}_{q_{\Theta}\left(\boldsymbol{z}_{i+1}\mid x_i,\boldsymbol{h}_{i-1}\right)}
\left[
\log p_{\Theta}\left(x_{i+1}\mid \boldsymbol{z}_{i+1}\right)
\right] \\
&+
\lambda
\operatorname{KL}
\left(
q_{\Theta}\left(\boldsymbol{z}_{i+1}\mid x_i,\boldsymbol{h}_{i-1}\right)
\|p\left(\boldsymbol{z}_{i+1}\right)
\right),
\end{aligned}
\end{equation}
where $p\left(\boldsymbol{z}_{i+1}\right)=\mathcal{N}(\boldsymbol{0},I)$, $\operatorname{KL}(\cdot)$ is the Kullback–Leibler (KL) divergence-based regularization, and $\lambda$ balances the likelihood term and latent-space regularization. 

In general, the variational sequence base model maps historical load information into a conditional predictive distribution for future loads. Depending on the forecasting requirement, the decoder can directly provide distributional parameters, such as the predictive mean and variance, or generate Monte Carlo samples by repeatedly drawing $\boldsymbol{z}_{i+1}$ and decoding the corresponding load realizations. These samples can further be used to estimate other probabilistic statistics, including quantiles and prediction intervals, making the base model flexible for both point and probabilistic forecasting. Such a model serves as the base model for M$^2$oE$^2$, as shown in the GRU encoder and mean/variance head blocks in Fig. \ref{fig:framework}.

\vspace{-2mm}
\subsection{External Meta-Knowledge through MetaNet and Gating}
\label{subsec:pre_meta}
After establishing the variational sequence base model, the next step is to incorporate external data in a way that helps the model adapt to changing environmental and contextual factors. Instead of simply concatenating external variables with load inputs, M$^2$oE$^2$ treats them as meta-knowledge that dynamically guides how the base model extracts load features.

For each time index $i$, optional external measurements are denoted as $\{w_{j,i}\mid j=1,\cdots,M\}$, where $M$ is the number of external data sources, such as weather, calendar variables, and electricity prices. Thus, we introduce the following modules to construct the final meta representation $\boldsymbol{\phi}_i$ in Eq. \ref{eqn:pre_moe}. 

\noindent\textbf{Module 1. MetaNet for building external experts.}
For the $j$-th external source, a MetaNet $g_j(\cdot)$ transforms $w_{j,i}$ into a candidate meta-representation:
\begin{equation}
\label{eqn:pre_hyper}
\boldsymbol{\psi}_{j,i}
=
\operatorname{LN}\left(g_j\left(w_{j,i}\right)\right),
\quad
j=1,\cdots,M,
\end{equation}
where $\operatorname{LN}(\cdot)$ denotes layer normalization. The layer normalization reduces scale and distributional heterogeneity among different external sources, so that each $\boldsymbol{\psi}_{j,i}$ can be treated as a candidate expert for guiding the forecasting model.

\noindent\textbf{Module 2. Gating network for selecting informative experts.}
Because the usefulness of external sources changes over time, M$^2$oE$^2$ employs a gating network $l(\cdot)$ to automatically select informative experts according to the historical hidden state. The combined meta-representation is
\begin{equation}
\label{eqn:pre_moe}
\boldsymbol{\phi}_i
=
\sum_{j=1}^{M} l_{j,i}\boldsymbol{\psi}_{j,i}
+
\boldsymbol{\phi}_0,
\end{equation}
where $l_{j,i}=l(\boldsymbol{h}_{i-1})[j]$ is the gating weight for the $j$-th external expert and $\boldsymbol{\phi}_0$ is a trainable residual anchor. A top-$m$ sparse softmax is applied to activate only the most relevant experts:
\begin{equation}
\label{eqn:pre_sparse_gate}
l_{j,i}
=
\frac{
\exp\left(\tilde{l}_{j,i}\right)\mathbb{I}\left(j\in\mathcal{T}_{i}\right)
}{
\sum_{r\in\mathcal{T}_{i}}\exp\left(\tilde{l}_{r,i}\right)
},
\end{equation}
where $\tilde{l}_{j,i}$ is the pre-activation score, $\mathcal{T}_{i}$ contains the indices of the selected top-$m$ experts, and $\mathbb{I}(\cdot)$ is the indicator function.

\begin{figure*}[t]
\centering
\includegraphics[
    width=\textwidth,
    trim={6mm 18mm 5mm 5mm},
    clip
]{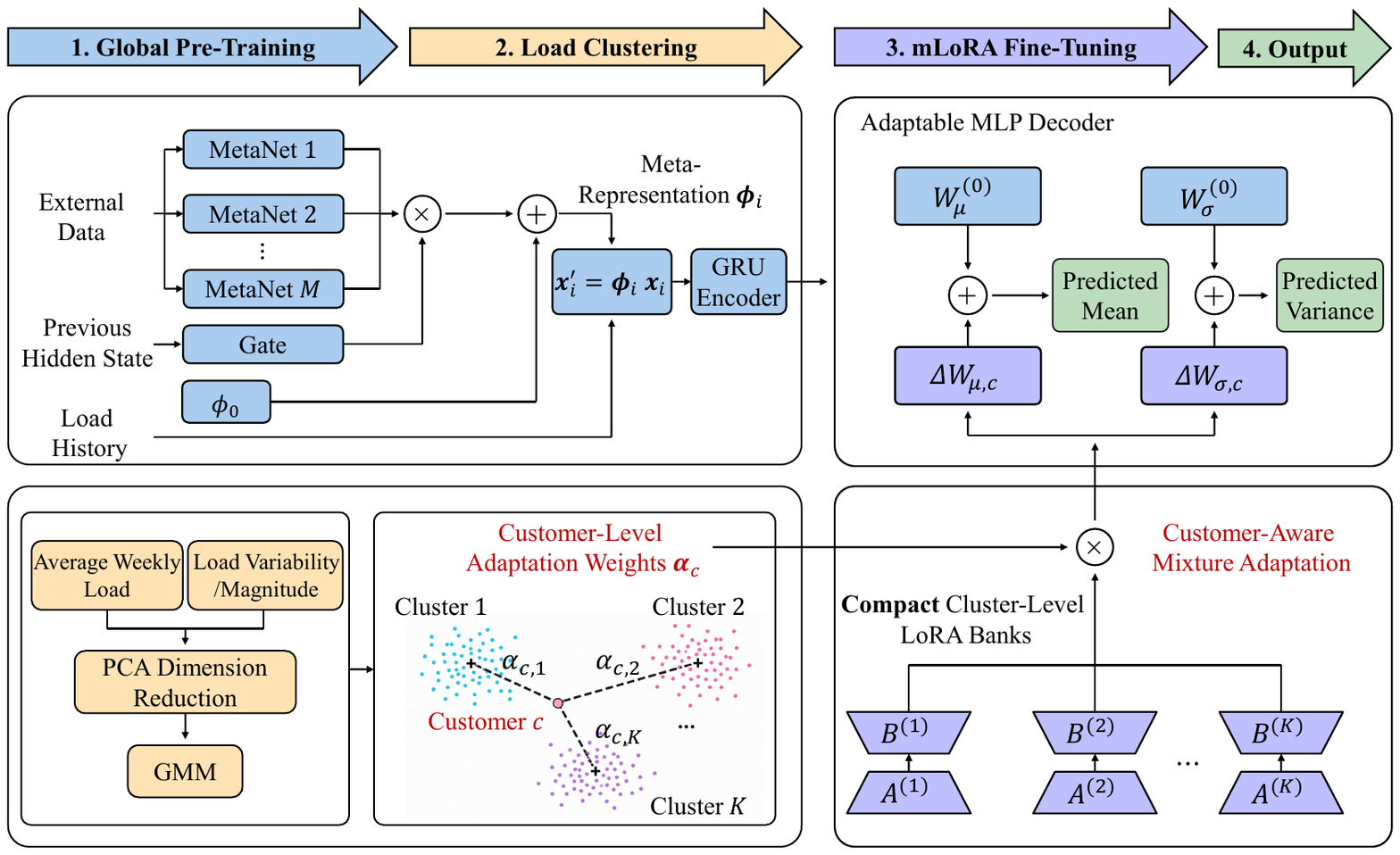}
\caption{M$^2$oE$^2$-mLoRA framework.}
\label{fig:framework}
\vspace{-4mm}
\end{figure*}

\noindent\textbf{Module 3. Skip connection for preserving the no-external-data case.}
The residual anchor $\boldsymbol{\phi}_0$ allows M$^2$oE$^2$ to fall back to the base forecasting structure when external information is uninformative. Specifically, if the selected external contribution vanishes, i.e., $\sum_{j=1}^{M} l_{j,i}\boldsymbol{\psi}_{j,i}=\boldsymbol{0}$, then $\boldsymbol{\phi}_i=\boldsymbol{\phi}_0$. Therefore, the external-data module can be bypassed, preventing the model from being worse than the corresponding no-external-data base model. The MetaNets, gate, and skip-connection are shown in the top-left part of Fig. \ref{fig:framework}. 

\vspace{-3mm}
\subsection{Plug-and-Play Parameter Modulation}
\vspace{-1mm}

\label{subsec:pre_plugin}

The meta-representation $\boldsymbol{\phi}_i$ can be used to modulate different sub-parameters of the sequence base model in Subsection \ref{subsec:pre_base}, such as input-to-hidden $W_{hx}$, hidden-to-hidden $W_{hh}$, or output-layer parameters in $\mu_x$ and $\sigma_x$. In M$^2$oE$^2$, a detailed analysis of parameter size, implementation cost, and adaptation effect shows that modulating the input-to-hidden matrix $W_{hx}$ is an efficient choice: it allows external conditions to change how load inputs are encoded into temporal features while avoiding the large overhead of modulating recurrent or output parameters.

Directly replacing the internal matrix $W_{hx}$ in Eq.~\eqref{eqn:pre_rnn}, however, requires modifying the internal implementation of the sequence base model. To obtain a plug-and-play design, M$^2$oE$^2$ uses an equivalent input transformation:
\begin{equation}
\label{eqn:pre_xprime}
\boldsymbol{x}_i^{\prime}
=
\boldsymbol{\phi}_i x_i.
\end{equation}
The transformed input $\boldsymbol{x}_i^{\prime}$, as shown on the top left of Fig. \ref{fig:framework}, is then fed into the sequence base model. This formulation enables external meta-knowledge to adapt the feature extraction without changing the internal architecture of the base model. As a result, M$^2$oE$^2$ can be integrated with different sequence base models, including RNN, GRU, and LSTM, and serves as an efficient global probabilistic forecasting backbone $f_{\Theta}$. Here, $\Theta=\Theta_{\mathrm{base}}\cup\Theta_{\mathrm{meta}}\cup\Theta_{\mathrm{gate}}\cup\{\boldsymbol{\phi}_0\}$ includes the parameters of the variational sequence base model, the MetaNets $\{g_j(\cdot)\}_{j=1}^{M}$, the gating network $l(\cdot)$, and the residual anchor $\boldsymbol{\phi}_0$.

In this paper, we further extend this backbone from external-data-aware forecasting to customer-aware forecasting by adapting sub-parameters $\{\theta_k\}_{k=1}^{K}$ for different load groups $k\in\{1,2,\cdots,K\}$, defined in Section \ref{sec:prob}.

\vspace{-3mm}

\section{Proposed Model}
\label{sec:model}
\vspace{-1mm}

In this section, we propose M$^2$oE$^2$-mLoRA, as shown in Fig.~\ref{fig:framework}. In the upper-left module (Stage 1), load history and external information are used to train a shared probabilistic forecaster; the pretrained backbone is then fixed during adaptation. In the lower-left module (Stage 2), load-profile features are clustered to identify representative customer and transformer patterns and determine the mixture weights used for adaptation. In the right module (Stage 3), these weights combine a shared bank of LoRA components to form customer-specific adaptations, which are applied to the probabilistic output heads for mean and variance prediction (Stage 4). This design reuses the same pretrained backbone across all loads while adapting only a compact set of parameters, avoiding full-model fine-tuning and supporting scalable deployment.

\vspace{-3mm}

\subsection{Load Clustering for Customer-Specific Fine-Tuning}
\label{subsec:load_clustering}

\vspace{-1mm}

Although the globally pre-trained M$^2$oE$^2$ backbone captures common temporal and external-context knowledge across massive loads, it may not fully address the strong heterogeneity among residential, commercial, industrial, and transformer-level loads. Therefore, after global pre-training, customer-aware adaptation is required to further specialize the model to different load behaviors by re-updating selected sub-parameters using specific groups of load data. Before performing such fine-tuning, it is necessary to understand how the load data are distributed across different behavioral patterns. To this end, we first conduct load clustering based on profile characteristics, so that loads with similar consumption patterns can provide structured guidance for the subsequent mixture-of-LoRA adaptation.

For each customer or transformer $c\in\mathcal{C}$, we use its historical training load data to construct one clustering representation. Specifically, the training data of load $c$ are divided into $Q$ weekly profiles, where each weekly profile is denoted by $\boldsymbol{x}_{c,q}\in\mathbb{R}^{S}$ ($1\leq q\leq Q$), and $S$ is the number of time slots in one week. Since clustering is expected to capture the load-shape characteristics rather than the absolute load magnitude, each weekly profile is normalized by the maximum training-period load of customer $c$. Specifically, we have: $x_c^{\max}
=
\max_{1\leq q\leq Q,\;1\leq s\leq S}
x_{c,q,s},
\tilde{\boldsymbol{x}}_{c,q}
=
\frac{\boldsymbol{x}_{c,q}}{x_c^{\max}+\epsilon},$ where $\epsilon$ is a small constant for numerical stability.

Then, all normalized weekly profiles are averaged to obtain a representative weekly load shape $\boldsymbol{p}_c$ for each customer or transformer. To characterize load heterogeneity beyond the average shape, we further consider week-to-week variability and load magnitude. Specifically, we define
$\boldsymbol{p}_c=\frac{1}{Q}\sum_{q=1}^{Q}\tilde{\boldsymbol{x}}_{c,q}$
and
$v_{c,s}=\left[\frac{1}{Q}\sum_{q=1}^{Q}\left(\tilde{x}_{c,q,s}-p_{c,s}\right)^2\right]^{1/2}$.
Here, $\boldsymbol{p}_c$ captures the average normalized weekly load shape, while $\boldsymbol{v}_c=[v_{c,1},\ldots,v_{c,S}]^\top$ characterizes the variability of each time slot across weeks. To retain absolute demand information removed by normalization, we further define
$\boldsymbol{a}_c=\left[\frac{1}{QS}\sum_{q=1}^{Q}\sum_{s=1}^{S}x_{c,q,s},\,x_c^{\max}\right]^\top$,
which contains the mean and maximum load magnitudes of the original profiles. The resulting feature vector therefore jointly characterizes typical load shape, temporal variability, and demand magnitude, corresponding to the three load-profile feature components in Fig.~\ref{fig:framework}:
\begin{equation}
\label{eqn:pc_build}
\boldsymbol{f}_c=
\left[
\boldsymbol{p}_c^{\top},
\boldsymbol{v}_c^{\top},
\boldsymbol{a}_c^{\top}
\right]^{\top}.
\end{equation}

Finally, the feature dimensions are standardized across the training loads and reduced using principal component analysis (PCA). The resulting low-dimensional representation $\boldsymbol{u}_c$ is used for subsequent clustering.

Given the reduced profile representations $\{\boldsymbol{u}_c\}_{c\in\mathcal{C}}$, we further apply a Gaussian mixture model (GMM) to identify representative load groups. The use of GMM is motivated by the fact that load groups in realistic distribution feeders may not have clear hard boundaries. In particular, transformer-level loads can aggregate multiple types of customers, such as residential, commercial, and industrial loads, leading to mixed profile characteristics. In addition, customer behavior may vary under different weather conditions, calendar effects, and operating regimes. Therefore, instead of assigning each load to a single deterministic cluster, GMM provides soft membership weights that describe how strongly each customer or transformer is associated with different load clusters.

\begin{equation}
\label{eqn:gmm_profile}
p(\boldsymbol{u}_c)
=
\sum_{k=1}^{K}
\omega_k
\mathcal{N}
\left(
\boldsymbol{u}_c
\mid
\boldsymbol{m}_k,
\Sigma_k
\right),
\end{equation}
where $K$ is the number of load groups, defined in Section \ref{sec:prob}. $\omega_k$ is the mixture coefficient satisfying $\sum_{k=1}^{K}\omega_k=1$, and $\boldsymbol{m}_k$ and $\Sigma_k$ are the mean vector and covariance matrix of the $k$-th Gaussian component. The soft membership weight of load $c$ with respect to cluster $k$ is given by the posterior probability
\begin{equation}
\label{eqn:gmm_posterior}
\alpha^{0}_{c,k}
=
\frac{
\omega_k
\mathcal{N}
\left(
\boldsymbol{u}_c
\mid
\boldsymbol{m}_k,
\Sigma_k
\right)
}{
\sum_{r=1}^{K}
\omega_r
\mathcal{N}
\left(
\boldsymbol{u}_c
\mid
\boldsymbol{m}_r,
\Sigma_r
\right)
},
\quad
k\in \{1,\cdots,K\}.
\end{equation}
The resulting vector $\boldsymbol{\alpha}^{0}_{c}
=
[\alpha^{0}_{c,1},\alpha^{0}_{c,2},\cdots,\alpha^{0}_{c,K}]$ serves as the clustering-based soft membership assignment of customer $c$ over the $K$ load groups. This profile construction, dimensionality reduction, and soft-membership estimation correspond to the load-profile features, PCA, and GMM blocks in the lower-left part of Fig.~\ref{fig:framework}.

\vspace{-3mm}
\subsection{Efficient Selection of Adapted Sub-Parameters}
\label{subsec:adapted_subparameters}
\vspace{-1mm}

The clustering provides a structured description of load distributions across different customers. However, these clustering assignments only suggest which data subsets may require distinct customer-specific adaptation; they do not specify which sub-parameters of M$^2$oE$^2$ should be adapted. A direct solution is to fine-tune the whole pre-trained model for each load cluster, but this would introduce large storage and computation overhead and weaken the scalability advantage of the shared backbone. Therefore, before constructing the mixture-of-LoRA adaptation, we first identify a small but effective subset of M$^2$oE$^2$ parameters for customer-specific fine-tuning.

For customer-aware forecasting, different sub-parameters of M$^2$oE$^2$ play different roles. The input transformation, external-data gating, and sequence feature extraction modules mainly learn common temporal and contextual representations shared by different loads. Directly adapting these modules for each customer group may disturb the globally learned representation and introduce unnecessary storage and computation overhead. In contrast, the \textbf{probabilistic output head} directly maps the learned latent representation to the predictive mean and variance, which are more sensitive to customer-specific load level, volatility, and uncertainty characteristics. Therefore, we select the output-head parameters as the adaptation target, so that each load group can adjust its predictive distribution while preserving the shared M$^2$oE$^2$ backbone.

Following Eq.~\eqref{eqn:pre_decoder}, for each customer or transformer $c\in\mathcal{C}$, the frozen M$^2$oE$^2$ backbone first produces the latent representation $\boldsymbol{z}_{c,i+1}$. The probabilistic output head then maps $\boldsymbol{z}_{c,i+1}$ to the mean and variance of the future load distribution:
\begin{equation}
\label{eqn:output_head}
\begin{aligned}
p_{\Theta}\left(x_{c,i+1}\mid \boldsymbol{z}_{c,i+1}\right)
&=
\mathcal{N}
\left(
\mu_x\left(\boldsymbol{z}_{c,i+1}\right),
\sigma_x^2\left(\boldsymbol{z}_{c,i+1}\right)
\right),\\
\mu_x\left(\boldsymbol{z}_{c,i+1}\right)
&=
\rho
\left(
W_{\mu}^{(0)}\boldsymbol{z}_{c,i+1}
+
\boldsymbol{b}_{\mu}^{(0)}
\right),\\
\log \sigma_x^2\left(\boldsymbol{z}_{c,i+1}\right)
&=
\rho
\left(
W_{\sigma}^{(0)}\boldsymbol{z}_{c,i+1}
+
\boldsymbol{b}_{\sigma}^{(0)}
\right),
\end{aligned}
\end{equation}
where $W_{\mu}^{(0)}$ and $W_{\sigma}^{(0)}$ are the pre-trained output-head weight matrices for predicting the mean and variance, respectively, and $\boldsymbol{b}_{\mu}^{(0)}$ and $\boldsymbol{b}_{\sigma}^{(0)}$ are the corresponding bias terms. Here, the superscript $(0)$ is added to $W_{\mu}^{(0)}$ and $W_{\sigma}^{(0)}$ to denote the globally pre-trained output-head weights that remain frozen during customer-specific adaptation. We select $W_{\mu}^{(0)}$ and $W_{\sigma}^{(0)}$ as the target sub-parameters for customer-specific adaptation, as shown in the upper-right module of Fig.~\ref{fig:framework}.

\vspace{-3mm}
\subsection{Soft-Membership-Guided Mixture-of-LoRA}
\vspace{-1mm}

\label{subsec:membership_mlora}

Subsequently, we need to determine how these parameters should be adapted for different customers and transformers. A straightforward strategy is to learn one independent LoRA update for each load group. However, this hard assignment ignores the soft membership information obtained from GMM and may be insufficient for transformer-level loads with mixed load profiles. Therefore, in this subsection, we propose a soft-membership-guided mixture-of-LoRA mechanism.

The basis of the proposed mixture-of-LoRA mechanism is a single LoRA update for one load cluster. Therefore, we first define how LoRA adapts the selected output-head parameters for the $k$-th cluster. Instead of updating the full pre-trained matrices $W_{\mu}^{(0)}$ and $W_{\sigma}^{(0)}$, LoRA represents the cluster-specific updates by low-rank decompositions:
\begin{equation}
\label{eqn:single_lora}
\begin{aligned}
W_{\mu}^{(k)}
&=
W_{\mu}^{(0)}
+
\Delta W_{\mu}^{(k)},
\quad
\Delta W_{\mu}^{(k)}
=
B_{\mu}^{(k)}A_{\mu}^{(k)},\\
W_{\sigma}^{(k)}
&=
W_{\sigma}^{(0)}
+
\Delta W_{\sigma}^{(k)},
\quad
\Delta W_{\sigma}^{(k)}
=
B_{\sigma}^{(k)}A_{\sigma}^{(k)},
\end{aligned}
\end{equation}
where $A_{\mu}^{(k)}\in\mathbb{R}^{r\times d_z}$, $B_{\mu}^{(k)}\in\mathbb{R}^{d_x\times r}$, $A_{\sigma}^{(k)}\in\mathbb{R}^{r\times d_z}$, and $B_{\sigma}^{(k)}\in\mathbb{R}^{d_x\times r}$ are the trainable low-rank factors for the $k$-th cluster, with $r\leq \min(d_z,d_x)$. Here, $d_z$ is the dimension of the latent representation $\boldsymbol{z}_{c,i+1}$, and $d_x$ is the output dimension of the forecasted load. In multi-step forecasting, $d_x$ is usually larger than $1$ because the output head predicts a future horizon; for example, $d_x=24$ for day-ahead hourly forecasting. During adaptation, the pre-trained matrices $W_{\mu}^{(0)}$ and $W_{\sigma}^{(0)}$ remain frozen, and only the low-rank factors are updated by gradients computed from the ELBO loss (see Eq. \eqref{eqn:pre_elbo}) on the load data associated with the $k$-th cluster. Such a mixture of LoRA banks is presented on the bottom right of Fig. \ref{fig:framework}.

As discussed in Subsection~\ref{subsec:load_clustering}, realistic load data often exhibit soft cluster memberships. Therefore, instead of selecting a single cluster-specific LoRA update, we construct a customer-specific adaptation as a linear mixture of multiple cluster-level LoRA updates. Let $\alpha_{c,k}$ denote the mixture weight of customer or transformer $c$ on the $k$-th cluster, where $\alpha_{c,k}\geq 0$ and $\sum_{k=1}^{K}\alpha_{c,k}=1$. The mixture-of-LoRA update is defined as:
\begin{equation}
\label{eqn:mlora_update}
\begin{aligned}
\Delta W_{\mu,c}
&=
\sum_{k=1}^{K}
\alpha_{c,k}
\Delta W_{\mu}^{(k)}
=
\sum_{k=1}^{K}
\alpha_{c,k}
B_{\mu}^{(k)}A_{\mu}^{(k)},\\
\Delta W_{\sigma,c}
&=
\sum_{k=1}^{K}
\alpha_{c,k}
\Delta W_{\sigma}^{(k)}
=
\sum_{k=1}^{K}
\alpha_{c,k}
B_{\sigma}^{(k)}A_{\sigma}^{(k)}.
\end{aligned}
\end{equation}

The adapted output-head matrices for customer or transformer $c$ are then given by
\begin{equation}
\label{eqn:mlora_head}
\begin{aligned}
W_{\mu,c}
&=
W_{\mu}^{(0)}
+
\Delta W_{\mu,c},\\
W_{\sigma,c}
&=
W_{\sigma}^{(0)}
+
\Delta W_{\sigma,c}.
\end{aligned}
\end{equation}

Eq.~\eqref{eqn:single_lora} studies the LoRA update for a single load group, where the $k$-th cluster has its own adapted output-head matrices $W_{\mu}^{(k)}$ and $W_{\sigma}^{(k)}$. In contrast, Eq.~\eqref{eqn:mlora_head} focuses on a specific customer $c$, which may have soft memberships to multiple load clusters rather than belonging to only one group. Therefore, the customer-specific output heads $W_{\mu,c}$ and $W_{\sigma,c}$ are constructed by combining the LoRA updates from different clusters according to the membership weights $\{\alpha_{c,k}\}_{k=1}^{K}$.

The remaining question is how to determine the mixture weight $\alpha_{c,k}$ in Eq.~\eqref{eqn:mlora_update}. The GMM posterior $\alpha^{0}_{c,k}$ from Eq.~\eqref{eqn:gmm_posterior} provides a useful intuition: for example, if customer or transformer $c$ is very close to the first load cluster, then the first entry of $\boldsymbol{\alpha}^{0}_{c}$ will be close to $1$ while the other entries will be close to $0$. In this case, Eq.~\eqref{eqn:mlora_update} should mainly activate the first LoRA update and suppress the others. Therefore, a direct option is to set $\alpha_{c,k}=\alpha^{0}_{c,k}$. 

However, $\alpha^{0}_{c,k}$ is computed only from the load-profile clustering space, while the optimal LoRA contribution is determined by the forecasting loss in the parameter-adaptation space. These two spaces are related but not identical. Thus, directly fixing $\alpha_{c,k}$ as $\alpha^{0}_{c,k}$ may overly restrict the mixture. To preserve the clustering intuition while allowing forecasting loss-based adjustment, we treat $\boldsymbol{\alpha}^{0}_{c}$ as the initialization of $\boldsymbol{\alpha}_{c}$ and introduce a trainable correction vector $\boldsymbol{r}_{c}=[r_{c,1},\cdots,r_{c,K}]$. Consequently, we can define:
\begin{equation}
\label{eqn:alpha_softmax}
\boldsymbol{\alpha}_{c}
=
\operatorname{softmax}
\left(
\log(\boldsymbol{\alpha}^{0}_{c}+\epsilon)
+
\tau \tanh(\boldsymbol{r}_{c})
\right),
\end{equation}
where $\epsilon$ is a small constant for numerical stability, and $\tau\geq 0$ controls the maximum deviation from the GMM-based soft membership assignment. The $\log$ operator converts the initial membership vector $\boldsymbol{\alpha}^{0}_{c}$ from probability space to score space, so that applying softmax with $\boldsymbol{r}_{c}=\boldsymbol{0}$ recovers $\boldsymbol{\alpha}^{0}_{c}$. Meanwhile, since $\tanh(\boldsymbol{r}_{c})$ is bounded within $[-1,1]$, the correction term cannot arbitrarily dominate $\log(\boldsymbol{\alpha}^{0}_{c}+\epsilon)$. Therefore, the mixture starts from the clustering-based membership assignment and remains guided by $\boldsymbol{\alpha}^{0}_{c}$ during training, while still allowing forecasting-loss-driven adjustment through the learnable correction vector $\boldsymbol{r}_{c}$.

We generate the correction score $\boldsymbol{r}_{c}$ using a shared lightweight neural network. For each customer-cluster pair $(c,k)$, the correction score is computed from the reduced customer profile $\boldsymbol{u}_c$ and the cluster center $\boldsymbol{m}_k$, defined in Eq. \eqref{eqn:gmm_posterior}. Specifically, we have:
\begin{equation}
\label{eqn:rc_network}
r_{c,k}
=
g_{\Omega}
\left(
[\boldsymbol{u}_c;\boldsymbol{m}_k]
\right),
\quad
k\in \{1,\cdots,K\},
\end{equation}
where $g_{\Omega}(\cdot)$ is a small MLP shared by all clusters. The resulting correction vector is $\boldsymbol{r}_c=[r_{c,1},\cdots,r_{c,K}]$. 

\begin{table*}[t]
\centering
\caption{Parameter-efficiency comparison of different customer-specific adaptation strategies.}
\label{tab:complexity_comparison}
\resizebox{\textwidth}{!}{
\begin{tabular}{lccc}
\toprule
Method 
& Required Parameters for Adaptation
& Complexity 
& Scaling w.r.t. $|\mathcal{C}|$ \\
\midrule

Full output-head adaptation
& 
$\{W_{\mu,c},W_{\sigma,c}\}_{c\in\mathcal{C}}$
&
$\mathcal{O}\!\left(2|\mathcal{C}|d_xd_z\right)$
&
Linear in $|\mathcal{C}|$
\\

Independent LoRA per customer
&
$\{A_{\mu,c},B_{\mu,c},A_{\sigma,c},B_{\sigma,c}\}_{c\in\mathcal{C}}$
&
$\mathcal{O}\!\left(2|\mathcal{C}|r(d_x+d_z)\right)$
&
Linear in $|\mathcal{C}|$
\\

Proposed M$^2$oE$^2$-mLoRA
&
$\{A_{\mu}^{(k)},B_{\mu}^{(k)},A_{\sigma}^{(k)},B_{\sigma}^{(k)}\}_{k=1}^{K}$, $g_{\Omega}$, PCA, GMM
&
\makecell{
$\mathcal{O}\!\left(2Kr(d_x+d_z)\right)$ \\
$+\mathcal{O}\!\left(2d_u+1\right)$ \\
$+\mathcal{O}\!\left(Sd_u+S\right)$ \\
$+\mathcal{O}\!\left(Kd_u+\frac{K d_u(d_u+1)}{2}+K\right)$
}
&
Independent of $|\mathcal{C}|$
\\

\bottomrule
\end{tabular}
}
\end{table*}

\begin{algorithm}[t]
\caption{Offline Training of M$^2$oE$^2$-mLoRA}
\label{alg:mlora_training}
\begin{algorithmic}[1]
\REQUIRE Training load data $\{\{x_{c,i}\}_{i=1}^{N_c}\}_{c\in\mathcal{C}}$, external data $\{w_{j,i}\}_{j=1}^{M}$, number of clusters $K$, LoRA rank $r$, and correction bound $\tau$.
\ENSURE Frozen backbone $\Theta^{(0)}$,
load features and statistics
$\boldsymbol{f}_c$, GMM parameters $\{\omega_k,\boldsymbol{m}_k,\Sigma_k\}_{k=1}^{K}$, mixture-of-LoRA factors $\Phi_{\mathrm{LoRA}}$,
and parameters $\Omega$ in $g_{\Omega}(\cdot)$.

\STATE \textbf{Step 1: Global pre-training}
\STATE Train the M$^2$oE$^2$ backbone using all loads by minimizing the ELBO objective in Eq.~\eqref{eqn:pre_elbo}.
\STATE Obtain the pretrained M$^2$oE$^2$, i.e., $f_{\Theta^{(0)}}$, and freeze $\Theta^{(0)}$.

\STATE \textbf{Step 2: Load clustering}
\FOR{each customer or transformer $c\in\mathcal{C}$}
    \STATE Construct the load-profile feature vector
    $\boldsymbol{f}_c$ by Eq.~\eqref{eqn:pc_build}.
\ENDFOR
\STATE Conduct PCA dimension reduction and obtain $\boldsymbol{u}_c$.
\STATE Fit a GMM with $K$ clusters on
$\{\boldsymbol{u}_c\}_{c\in\mathcal{C}}$
using Eq.~\eqref{eqn:gmm_profile}.
\STATE Obtain the soft membership assignment
$\boldsymbol{\alpha}^{0}_{c}$
by Eq.~\eqref{eqn:gmm_posterior}.

\STATE \textbf{Step 3: Cluster-wise LoRA pretraining}
\FOR{each load cluster $k=1,\cdots,K$}
    \STATE Pretrain the single-cluster LoRA basis in Eq.~\eqref{eqn:single_lora} by minimizing the weighted ELBO loss in Eq.~\eqref{eqn:lora_warm_start}.
    \STATE Obtain the pretrained LoRA basis $\Phi_{\mathrm{LoRA}}^{(k),\mathrm{init}}$.
\ENDFOR
\STATE Set $\Phi_{\mathrm{LoRA}}
\leftarrow
\left\{
\Phi_{\mathrm{LoRA}}^{(k),\mathrm{init}}
\right\}_{k=1}^{K}$.
\vspace{1mm}
\STATE \textbf{Step 4: Joint mixture-of-LoRA adaptation}
\WHILE{not converged}
    \STATE Sample a mini-batch of customer-time pairs $\mathcal{B}$.
    \FOR{each sampled customer or transformer $c$ in $\mathcal{B}$}
        \STATE Compute the correction vector $\boldsymbol{r}_c$ by Eq.~\eqref{eqn:rc_network}.
        \STATE Compute the bounded corrected mixture weight $\boldsymbol{\alpha}_{c}$ by Eq.~\eqref{eqn:alpha_softmax}.
        \STATE Construct $W_{\mu,c}$ and $W_{\sigma,c}$ by Eqs.~\eqref{eqn:mlora_update} and~\eqref{eqn:mlora_head}.
    \ENDFOR
    \STATE Obtain the predictive mean and variance by replacing $W_{\mu}^{(0)}$ and $W_{\sigma}^{(0)}$ in Eq.~\eqref{eqn:output_head} with the customer-specific adapted matrices $W_{\mu,c}$ and $W_{\sigma,c}$.
    \STATE Compute the ELBO-based loss in Eq.~\eqref{eqn:pre_elbo} on the sampled mini-batch $\mathcal{B}$.
    \STATE Update only $\Phi_{\mathrm{LoRA}}$ and $\Omega$ in $g_{\Omega}(\cdot)$ by backpropagation.
\ENDWHILE

\RETURN $\Theta^{(0)}$, $\{\omega_k,\boldsymbol{m}_k,\Sigma_k\}_{k=1}^{K}$, $\Phi_{\mathrm{LoRA}}$, and $\Omega$.

\end{algorithmic}
\end{algorithm}

\vspace{-3mm}

\subsection{Training Algorithm and Online Prediction}
\label{subsec:training_prediction}
\vspace{-1mm}

In this subsection, we describe the offline training and online prediction procedures of the proposed M$^2$oE$^2$-mLoRA. The offline training procedure, summarized in Algorithm~\ref{alg:mlora_training}, consists of four stages: global pre-training, load clustering, cluster-wise LoRA pretraining, and joint mixture-of-LoRA adaptation. First, the M$^2$oE$^2$ backbone is trained on all available loads by minimizing the ELBO objective in Eq.~\eqref{eqn:pre_elbo}, and the pretrained backbone is then frozen. Second, feature preparation, PCA, and GMM are applied to obtain the clustering-based soft membership assignment $\boldsymbol{\alpha}^{0}_{c}$. Third, each single-cluster LoRA basis in Eq.~\eqref{eqn:single_lora} is pretrained using the load data associated with its corresponding cluster. Since GMM provides soft memberships, the contribution of customer or transformer $c$ to the $k$-th LoRA pretraining loss is weighted by $\alpha^{0}_{c,k}$:
\begin{equation}
\label{eqn:lora_warm_start}
\begin{aligned}
\Phi_{\mathrm{LoRA}}^{(k),\mathrm{init}}
&=
\arg\min_{\Phi_{\mathrm{LoRA}}^{(k)}}
\sum_{c\in\mathcal{C}}
\alpha_{c,k}^{0} \\
&\qquad \times
\sum_{i=1}^{N_c}
\mathcal{L}_{\mathrm{ELBO}}\!\left(
x_{c,i+1};
W_{\mu}^{(k)},W_{\sigma}^{(k)}
\right),
\end{aligned}
\end{equation}
where $\Phi_{\mathrm{LoRA}}^{(k)}
=
\{A_{\mu}^{(k)},B_{\mu}^{(k)},A_{\sigma}^{(k)},B_{\sigma}^{(k)}\}.$ The resulting pretrained single-cluster LoRA bases are then used to initialize the mixture-of-LoRA bank: $\Phi_{\mathrm{LoRA}}
\leftarrow
\left\{
\Phi_{\mathrm{LoRA}}^{(k),\mathrm{init}}
\right\}_{k=1}^{K}.$ This warm-start step gives each LoRA basis an explicit cluster-level meaning before the final joint adaptation stage. After the warm-start, the cluster-level LoRA bases and the correction network $g_{\Omega}(\cdot)$ are jointly optimized while the M$^2$oE$^2$ backbone remains frozen. During joint adaptation, only $\Phi_{\mathrm{LoRA}}$ and $\Omega$ are updated by minimizing the ELBO-based loss in Eq.~\eqref{eqn:pre_elbo}, where the predictive mean and variance are computed using the customer-specific adapted output heads $W_{\mu,c}$ and $W_{\sigma,c}$.

During online prediction, the frozen M$^2$oE$^2$ backbone is reused for all customers. For each customer or transformer $c$, the trained PCA, GMM, correction network, and mixture-of-LoRA factors are used to construct the customer-specific adapted output heads $W_{\mu,c}$ and $W_{\sigma,c}$. The final predictive mean and variance are then obtained by applying Eq.~\eqref{eqn:output_head} with $W_{\mu}^{(0)}$ and $W_{\sigma}^{(0)}$ replaced by $W_{\mu,c}$ and $W_{\sigma,c}$.

\vspace{-3mm}

\subsection{Parameter Efficiency and Deployment Complexity}
\label{subsec:complexity}
\vspace{-1mm}

In realistic distribution feeders, $|\mathcal{C}|$ (the number of customers or transformers) can be very large because utilities may need to forecast thousands or even hundreds of thousands of customer- and transformer-level loads. Therefore, the storage complexity and scaling with respect to $|\mathcal{C}|$ becomes a critical deployment bottleneck. Table~\ref{tab:complexity_comparison} presents the parameter-efficiency comparison among three customer-specific adaptation strategies. Here, $d_z$ is the dimension of the latent representation, $d_x$ is the output dimension of the forecasting horizon, $r$ is the LoRA rank, $K$ is the number of load clusters, $S$ is the dimension of the normalized weekly profile, and $d_u$ is the PCA-reduced profile dimension.

For full output-head adaptation, each customer requires its own mean and variance output matrices, resulting in $\mathcal{O}(2|\mathcal{C}|d_xd_z)$ additional parameters. This cost grows linearly with the number of loads and becomes expensive in large-scale feeders. Independent LoRA per customer reduces the cost by replacing each full matrix update with low-rank factors. For each customer, the mean and variance heads require two LoRA updates, leading to $\mathcal{O}(2|\mathcal{C}|r(d_x+d_z))$ additional parameters. Although this is much smaller than full output-head adaptation when $r\ll \min(d_x,d_z)$, the storage cost still scales linearly with $|\mathcal{C}|$, since each customer or transformer needs an independent adapter.

In contrast, the proposed M$^2$oE$^2$-mLoRA stores only $K$ cluster-level LoRA updates shared by all customers, giving $\mathcal{O}(2Kr(d_x+d_z))$ parameters for the mean and variance heads. In addition, the lightweight (1 layer) neural network $g_{\Omega}$ maps $[\boldsymbol{u}_c;\boldsymbol{m}_k]\in\mathbb{R}^{2d_u}$ to the scalar correction score $r_{c,k}$, requiring $\mathcal{O}(2d_u+1)$ parameters. The clustering-related components introduce $\mathcal{O}(Sd_u+S)$ parameters for the PCA projection and mean profile, and $\mathcal{O}(Kd_u+\frac{Kd_u(d_u+1)}{2}+K)$ parameters for the full-covariance GMM. Since these components are shared across all customers and $K\ll |\mathcal{C}|$, the proposed method avoids storing customer-specific adapters and keeps the additional parameter cost independent of the number of loads.

\vspace{-3mm}

\begin{table*}[t]
\centering
\caption{Forecasting performance and computational cost.}
\label{tab:benchmark}

\footnotesize
\setlength{\tabcolsep}{2.5pt}
\renewcommand{\arraystretch}{1.10}

\begin{tabular*}{\textwidth}{@{\extracolsep{\fill}}lrrrrrrrrrr@{}}
\toprule
Model
& MSE ($10^{-3}$)
& RMSE
& MAE
& WAPE
& CRPS
& QS
& WS$_{90}$
& WS$_{50}$
& Time
& Storage \\
\midrule

ARIMA + Fourier
& 5.47 & 8.79 & 3.26 & 8.19
& 2.45 & 1.10 & 20.96 & 11.16
& 6.86$^{*}$ & \textbf{0.18} \\

CNN-GRU
& 3.92 & 6.58 & 2.37 & 5.95
& 1.75 & 0.79 & 15.52 & 7.79
& \textbf{0.04} & 0.33 \\

TimesFM-2.5 zero shot
& 3.74 & 7.22 & 2.35 & 5.91
& 1.80 & 0.83 & 17.72 & 8.03
& 4.92 & 882.30 \\

iTransformer
& 3.69 & 6.22 & 2.14 & 5.37
& 1.67 & 0.77 & 16.99 & 7.42
& 0.06 & 18.77 \\

LSTM
& 3.51 & 4.90 & 1.77 & 4.45
& 1.31 & 0.59 & 11.88 & 5.84
& 0.10 & 0.79 \\

GRU
& 3.49 & 5.44 & 1.98 & 4.97
& 1.48 & 0.68 & 14.26 & 6.57
& 0.06 & 0.60 \\

PatchTST
& 3.35 & 5.43 & 1.87 & 4.68
& 1.45 & 0.68 & 15.36 & 6.44
& \textbf{0.04} & 2.32 \\

Chronos-2 zero shot
& 3.21 & 6.16 & 2.02 & 5.07
& 1.48 & 0.67 & 12.59 & 6.61
& 10.15 & 455.77 \\

DeepAR
& 3.18 & 5.08 & 1.71 & 4.28
& 1.45 & 0.72 & 12.78 & 6.98
& 11.62 & 0.77 \\

TimesFM-2.5 + LoRA
& 3.15 & 5.32 & 1.73 & 4.35
& 1.37 & 0.64 & 14.99 & 6.07
& 4.93 & 882.30 \\

M$^2$oE$^2$
& 2.98 & 4.87 & 1.79 & 4.49
& 1.32 & 0.59 & 11.29 & 5.90
& 0.08 & 0.40 \\

M$^2$oE$^2$-mLoRA
& \textbf{2.72}
& \textbf{4.50}
& \textbf{1.60}
& \textbf{4.16}
& \textbf{1.18}
& \textbf{0.48}
& \textbf{11.01}
& \textbf{5.54}
& 0.11 & 0.58 \\

\bottomrule
\end{tabular*}

\par\smallskip
\begin{minipage}{\textwidth}
\footnotesize
MSE uses each customer's training load range and is reported in units of
$10^{-3}$. RMSE, MAE, CRPS, QS, WS$_{90}$, and WS$_{50}$ are in kW,
while WAPE is reported in percent. Time is reported in ms/forecast and
storage in MiB. Bold values indicate the best result in each column before
rounding. Accuracy is evaluated on the full December test set. Time is the
median processing time divided by 4,130 daily forecasts from December 3--9,
with each forecast covering 24 hours. Neural models are evaluated on an
NVIDIA Tesla T4. $^{*}$ARIMA is evaluated on an Intel Xeon CPU with two
configured threads and is excluded from GPU speed comparisons. Storage
includes the forecasting model and required static routing data, with one
MiB equal to $2^{20}$ bytes.
\end{minipage}
\end{table*}

\section{Experiments}
\label{sec:exp}

\subsection{Setup}
\label{subsec:set}
SMART-DS (Synthetic Models for Advanced, Realistic Testing: Distribution Systems and Scenarios) provides public distribution network models and load time series for several U.S. regions, including San Francisco, Greensboro, and Austin~\cite{smartds2020}. The models are designed to reproduce realistic statistical and operational characteristics of utility distribution systems, with active and reactive power profiles available at 15-minute intervals. We use 590 residential, commercial, and industrial load profiles from the 2018 dataset and aggregate active power to hourly resolution for forecasting. Selected profiles contain complete observations throughout 2018 and have a 95th-percentile hourly demand of at least 4~kW from January to October, spanning demand levels from 4~kW to above 128~kW.

Training uses January-October data, validation uses November, and testing uses December. Each day at midnight, the model forecasts the next 24 hours. Using complete weekly samples yields 30 forecasting times per profile from December 1–30, 2018, corresponding to 17,700 daily forecasts and 424,800 hourly predictions. Each sample spans two weeks. The model first encodes 168 hours of historical load, then updates this representation with newly available observations before each daily forecast and predicts the next 24 hours' load.

We evaluate forecasting performance using seven metrics \cite{li2025external,ref:haoran2026AF}: Mean Squared Error (MSE), Root Mean Squared Error (RMSE), Mean Absolute Error (MAE), Weighted Absolute Percentage Error (WAPE), Continuous Ranked Probability Score (CRPS), Quantile Score (QS), and Winkler Score (WS). MSE measures profile-normalized squared error; RMSE and MAE quantify prediction errors in kW; WAPE measures absolute error relative to aggregate demand; CRPS evaluates the overall predictive distribution; QS assesses the accuracy of predicted quantiles; and WS evaluates prediction intervals by jointly considering their coverage and width. Lower values indicate better performance for all metrics.


\begin{figure*}[!tp]
\centering
\includegraphics[
    width=0.96\textwidth,
    height=0.2\textheight,
    trim={2mm 2mm 2mm 2mm},
    clip
]{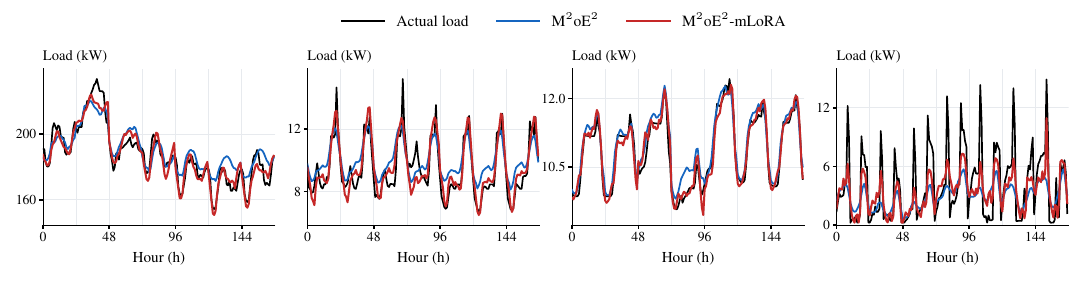}
\caption{Examples of load forecasts from M$^2$oE$^2$ and M$^2$oE$^2$-mLoRA.}
\label{fig:forecast_examples}
\vspace{-4mm}
\end{figure*}

We compare M$^2$oE$^2$-mLoRA against statistical, recurrent/convolutional, Transformer-based, and pretrained time-series foundation models. The benchmarks include ARIMA + Fourier, GRU~\cite{cho2014learning}, LSTM~\cite{hochreiter1997long}, CNN-GRU, DeepAR~\cite{salinas2020deepar}, iTransformer~\cite{liu2023itransformer}, PatchTST~\cite{nie2023patchtst}, Chronos-2~\cite{ansari2025chronos2}, TimesFM-2.5~\cite{das2024timesfm,timesfmrepo}, and TimesFM-2.5 + LoRA~\cite{hu2022lora}, covering deterministic and probabilistic forecasting as well as zero-shot and adapted foundation models. We further compare against an M$^2$oE$^2$ ablation~\cite{li2025external} without the proposed mLoRA adaptation, which serves as the shared base model for evaluating the contribution of customer-specific adapters.

\begin{figure*}[!tp]
\centering
\includegraphics[width=\textwidth]{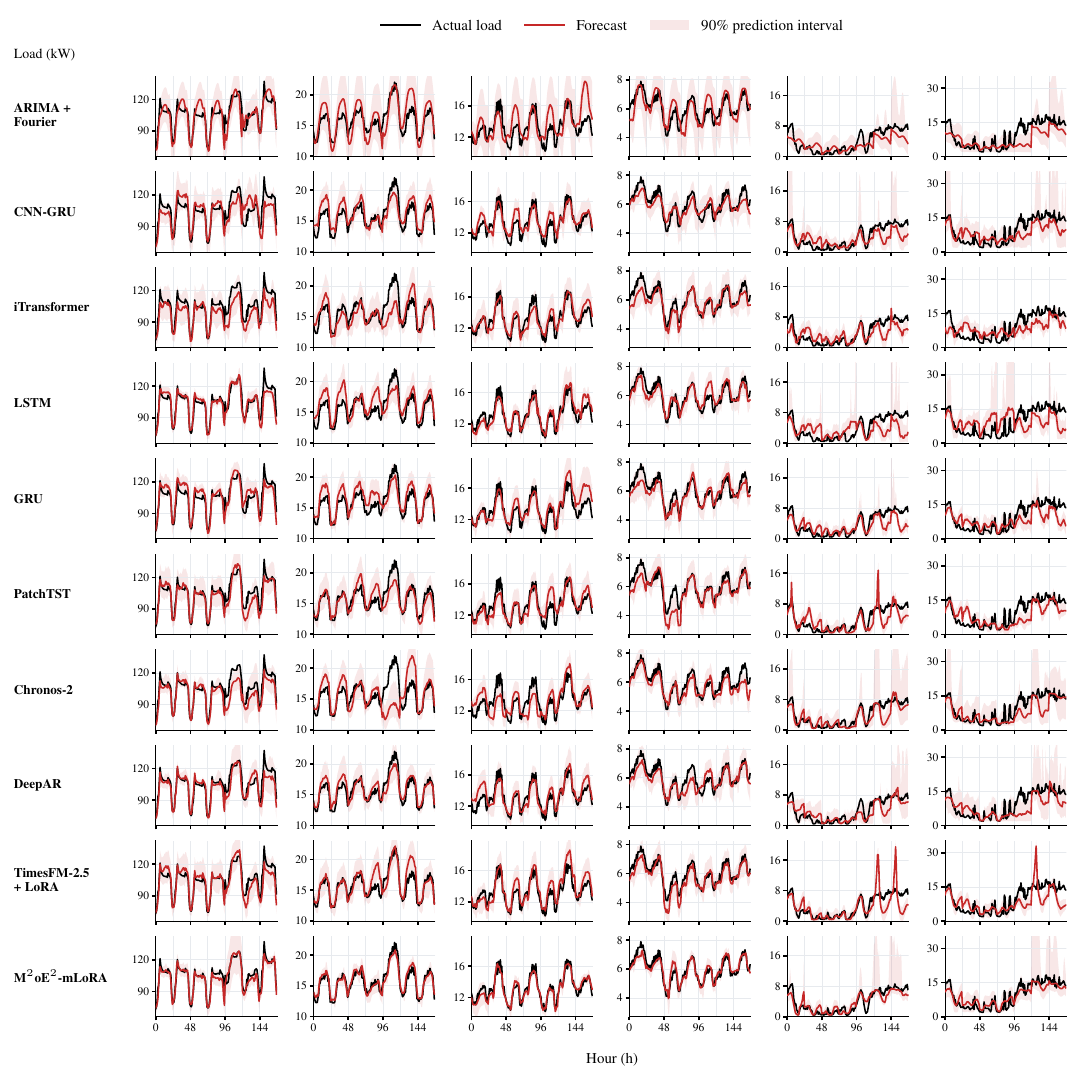}
\caption{Weekly load forecasts for six selected SMART-DS profiles. Rows show all nine benchmark configurations, and M$^2$oE$^2$-mLoRA. Each column uses the same actual loads and vertical scale throughout. }
\label{fig:weekly_benchmarks_a}
\label{fig:benchmark_4x4}
\vspace{-4mm}
\end{figure*}

\vspace{-3mm}
\subsection{Forecasting Performance and Computational Cost}
\vspace{-1mm}

Table~\ref{tab:benchmark} shows that M$^2$oE$^2$-mLoRA achieves the best performance across all evaluated deterministic and probabilistic forecasting metrics. Its MSE is 8.72\% lower than that of M$^2$oE$^2$ and 13.65\% lower than that of TimesFM-2.5 with LoRA, demonstrating the benefit of cluster-specific mLoRA adaptation. The following figures further illustrate the forecasting behavior for representative load profiles.

Fig.~\ref{fig:forecast_examples} compares the actual loads with forecasts from M$^2$oE$^2$ and M$^2$oE$^2$-mLoRA for representative profiles with different demand levels and temporal patterns. Each panel shows seven consecutive day-ahead forecasts. For the first two profiles, customer-aware adaptation reduces overestimation between daily peaks, while for the third profile, it more accurately captures the lower-load period around hours 72-90. The last profile exhibits more irregular fluctuations, for which M$^2$oE$^2$-mLoRA better tracks low-load intervals and more accurately recovers a sharp demand spike. Fig.~\ref{fig:weekly_benchmarks_a} compares one-week forecasts for six test profiles across all benchmark models. Overall, M$^2$oE$^2$-mLoRA tracks the observed load levels and daily variations more closely than the compared methods.

Inference cost is evaluated on the December 3-9 test period under the same forecasting setting for all methods. As shown in Table~\ref{tab:benchmark}, M$^2$oE$^2$-mLoRA requires only 0.11 ms per forecast and 0.58 MiB of storage, remaining substantially more efficient than large pretrained foundation models while adding only modest overhead over the shared M$^2$oE$^2$ base. This demonstrates that the proposed cluster-specific adaptation improves forecasting accuracy without sacrificing computational scalability.

\begin{table}
\centering
\footnotesize
\caption{Sensitivity to the number of clusters.}
\label{tab:cluster_sensitivity}
\begin{tabular}{@{}rrr@{}}
\toprule
$K$ & Adapter parameters & November MSE ($\times 10^{-3}$)\\
\midrule
14 & 1,971,200 & 2.97\\
15 & 2,112,000 & \textbf{2.96}\\
16 & 2,252,800 & 2.97\\
\bottomrule
\end{tabular}
\vspace{-3mm}
\end{table}

\vspace{-3mm}

\subsection{Parameter Sensitivity}
\vspace{-1mm}

We evaluate the sensitivity of M$^2$oE$^2$-mLoRA to the number of load clusters using the same 590 profiles and chronological data split. The clusters are constructed from training data based on similarities in training-period load-profile features, and each cluster is assigned its own mLoRA adapter. As $K$ increases, the model can represent more diverse customer-specific adaptation patterns, but the number of adapter parameters and overall model size also increase approximately linearly. Table~\ref{tab:cluster_sensitivity} compares $K=14,15,16$; the lowest November validation MSE is obtained with $K=15$, while the overall performance remains stable across the tested values, supporting the use of 15 clusters in the final model.
%
%

\vspace{-3mm}

\section{Conclusion}
\label{sec:con}

This paper studied scalable customer-aware probabilistic forecasting for heterogeneous customer- and transformer-level loads without training a separate full model for each load. The proposed M$^2$oE$^2$-mLoRA framework identifies the probabilistic output head as the key parameter subset for customer-specific adaptation and introduces a soft-membership-guided mixture-of-LoRA mechanism, allowing each load to combine compact cluster-level adaptations while preserving the shared forecasting backbone. Experiments on 590 load profiles from the SMART-DS dataset show that M$^2$oE$^2$-mLoRA improves all evaluated deterministic and probabilistic forecasting metrics over the shared M$^2$oE$^2$ model and achieves the best overall forecasting performance among the evaluated methods, with low additional storage and inference cost. These results demonstrate that mixture-based parameter-efficient adaptation provides a scalable approach to capturing heterogeneous demand and uncertainty patterns in customer- and transformer-level probabilistic forecasting.

\bibliographystyle{IEEEtran}
\bibliography{reference}
\end{document}